\documentclass{article}

\usepackage{PRIMEarxiv}
\usepackage{dirtytalk}
\usepackage[utf8]{inputenc} 
\usepackage[T1]{fontenc}    
\usepackage{hyperref}       
\usepackage{url}            
\usepackage{booktabs}       
\usepackage{amsfonts}       
\usepackage{nicefrac}       
\usepackage{microtype}      
\usepackage{lipsum}
\usepackage{fancyhdr}       
\usepackage{graphicx}       
\graphicspath{{media/}}     

\title{MemeFier: Dual-stage modality fusion for image meme classification
\thanks{\textit{\underline{Citation}}: 
\textbf{Koutlis, C., Schinas, M. \& Papadopoulos, S. (2023). MemeFier: Dual-stage modality fusion for image meme classification. Accepted at ICMR 2023 as a short paper.}}
}

\author{
  Christos Koutlis, Manos Schinas, Symeon Papadopoulos  \\
  CERTH-ITI \\
  Thessaloniki \\
  Greece\\
  \texttt{\{ckoutlis,manosetro,papadop\}@iti.gr} \\
}

\begin{document}
\maketitle

\begin{abstract}
Hate speech is a societal problem that has significantly grown through the Internet. New forms of digital content such as \textit{image memes} have given rise to spread of hate using multimodal means, being far more difficult to analyse and detect compared to the unimodal case. Accurate automatic processing, analysis and understanding of this kind of content will facilitate the endeavor of hindering hate speech proliferation through the digital world. To this end, we propose MemeFier, a deep learning-based architecture for fine-grained classification of Internet image memes, utilizing a dual-stage modality fusion module. The first fusion stage produces feature vectors containing modality alignment information that captures non-trivial connections between the text and image of a meme. The second fusion stage leverages the power of a Transformer encoder to learn inter-modality correlations at the token level and yield an informative representation. Additionally, we consider external knowledge as an additional input, and background image caption supervision as a regularizing component. Extensive experiments on three widely adopted benchmarks, i.e., Facebook Hateful Memes, Memotion7k and MultiOFF, indicate that our approach competes and in some cases surpasses state-of-the-art. Our code is available on GitHub\footnote{\url{https://github.com/ckoutlis/memefier}}.
\end{abstract}


\section{Introduction}
\begin{figure}
    \centering
    \includegraphics[width=0.6\textwidth]{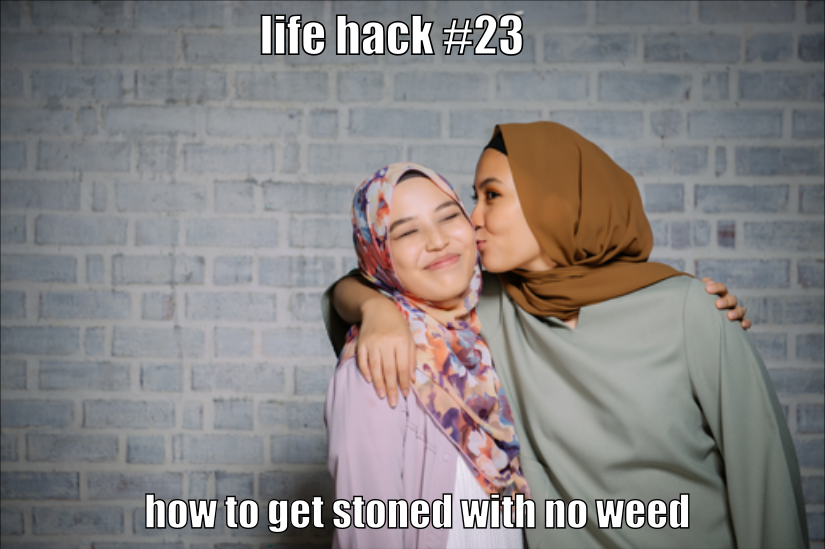}
    \caption{Hateful meme example from \cite{kiela2020hateful}.}
    \label{fig:hateful}
\end{figure}

One common way to express emotions and opinions in the digital world is through \textit{image memes}, whose primary intention is to be funny or motivational. However, memes have also been utilized as a means to offend, cyber-bully and in general spread hate (cf. Figure~\ref{fig:hateful}). This unfortunate fact along with the huge scale of daily uploads of such content, which renders human-only moderation impractical, stimulated the research community to devise mechanisms for automatic detection of image memes \cite{koutlis2022memetector}, recognition of the type of humour expressed by an image meme, detection of the existence and the level of offensiveness, and identification of the attack's target group \cite{sharma2020semeval, kiela2021hateful, mathias-etal-2021-findings}.

The under study problem of image meme classification has been proven to be challenging \cite{sharma2022detecting,chhabra2023literature,GANDHI2023424}.  One of the main reasons is its multimodal nature, given that the interpretation of the textual part of the meme often is tied to the visual part and vice versa. For processing and analysing such media, there is no meaning in understanding the overlay text or the background image in isolation because the expressed message is produced by the combination of the two modalities. This makes the problem hard to address in a fully automatic manner, as the  association between modalities is quite complex contrary to other multimodal tasks, where both modalities complement towards the same meaning, or in the case of text-only hate speech detection.

In order to alleviate the above problem we propose MemeFier, a deep learning-based architecture that utilizes a dual-stage modality fusion module to yield meme-oriented features. In the first modality fusion stage, we attempt to estimate the level of alignment between the two modalities by a simple element-wise multiplication of the pre-trained representations at the token level. Then, the alignment-aware features are processed by a Transformer encoder that estimates intra- and inter-modality associations with the multihead softmax attention mechanism, to produce a rich feature vector. Finally, we also consider external knowledge as input to our model as well as a caption supervision relevant to the background image as a regularizing factor.


\section{Related work}\label{sec:related_work}
Although the topic of meme classification has emerged  recently, there are already several papers addressing it and consecutively pushing further the state-of-the-art on pertinent standard benchmarks such as Memotion7k \cite{sharma2020semeval}, Facebook Hateful Memes \cite{kiela2021hateful} and MultiOFF \cite{suryawanshi2020multimodal}.

Exploratory data analysis methods, such as t-SNE \cite{van2008visualizing}, have been used for cluster detection \cite{bonheme-grzes-2020-sesam}, and Canonical Correlation Analysis (CCA) \cite{Yang21} and variants (e.g., Kernel CCA \cite{lai2000kernel}, Deep CCA \cite{pmlr-v28-andrew13}) for the study of modality interrelationship \cite{bonheme-grzes-2020-sesam}. Many modality fusion approaches have been utilized such as early fusion (embedding concatenation \cite{guo-etal-2020-nuaa,gundapu-mamidi-2020-gundapusunil,guo-etal-2020-guoym,Alluri21} or sum/average \cite{baruah-etal-2020-iiitg}, Gated Multimodal Unit \cite{gupta-etal-2020-dsc}, multihead softmax attention \cite{10.1007/978-3-030-99736-6_35}, Low-rank factorization bilinear pooling \cite{kumari-etal-2021-co} and outer product \cite{10.1007/978-3-030-98358-1_47,kumar-nandakumar-2022-hate}), late fusion \cite{gupta-etal-2020-dsc,shrestha-rusert-2020-nlp}, hybrid fusion \cite{bonheme-grzes-2020-sesam,yuan-etal-2020-ynu} and even inter-task fusion in cases of multi-task learning \cite{10.1007/978-3-030-99736-6_35}. Also, a typical approach is to consider pre-trained modality representations using word embeddings (e.g., GloVe, FastText, Word2Vec) \cite{keswani-etal-2020-iitk-semeval,kumari-etal-2021-co} and large-scale pre-trained neural networks for text \cite{10.1007/978-3-030-99736-6_35,Alluri21}, image \cite{Alluri21} and multimodal processing \cite{kiela2021hateful,zhu2020enhance,muennighoff2020vilio,velioglu2020detecting}.

Most of the previous works leverage ensemble approaches to achieve improved performance. The main model ensemble techniques considered are soft voting \cite{muennighoff2020vilio,shrestha-rusert-2020-nlp}, majority voting \cite{yuan-etal-2020-ynu} as well as other linear and non-linear \cite{muennighoff2020vilio} prediction models optimally combining the ensemble models' predictions. The ensemble members typically are (i) large-scale VL pre-trained models \cite{zhu2020enhance}, (ii) neural networks trained on k-folds of the training set \cite{lippe2020multimodal}, (iii) from scratch trained neural networks initialized with different seeds \cite{muennighoff2020vilio,sandulescu2020detecting} or of different hyperparameter settings \cite{velioglu2020detecting} and, (iv) produced by evolutionary algorithms \cite{lippe2020multimodal}.

Finally, external knowledge has been leveraged by studies to make the models aware of information not provided explicitly by the datasets. The most frequent choices of external knowledge are textual and visual sentiment predictions \cite{gupta-etal-2020-dsc}, protected attribute extraction \cite{zhu2020enhance,10.1145/3474085.3475625}, entity linking in the text \cite{10.1007/978-3-030-98358-1_47} and the image \cite{zhu2020enhance,10.1145/3474085.3475625}, and fine-grained object detection predictions \cite{velioglu2020detecting,lippe2020multimodal,10.1145/3474085.3475625}.

\section{MemeFier}\label{sec:memefier}
\begin{figure*}
    \centering
    \includegraphics[width=\textwidth]{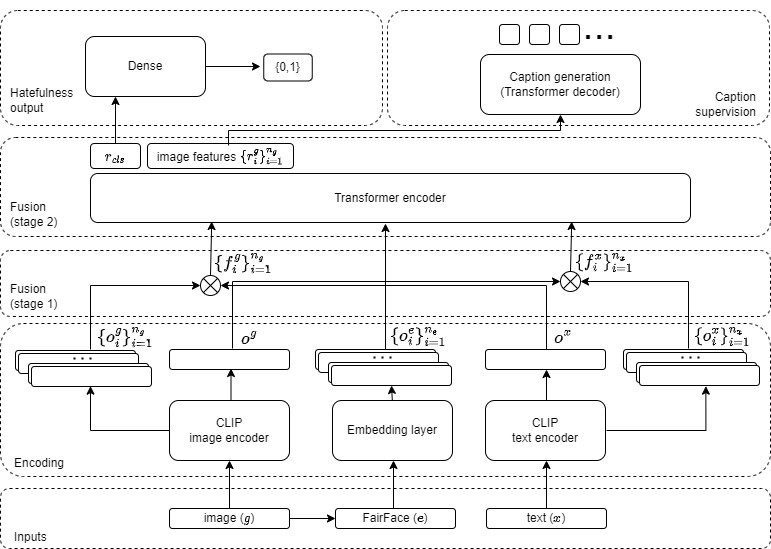}
    \caption{MemeFier's architecture. $\otimes$ denotes broadcastable element-wise multiplication.}
    \label{fig:architecture}
\end{figure*}

Here we present MemeFier, a method to address the image meme classification task. 
Figure~\ref{fig:architecture} illustrates the proposed architecture.

\subsection{Modality encoding}
In \cite{kiela2021hateful}, it is claimed that feature extractors trained with a multimodal objective tend to outperform combined unimodally pre-trained ones, when the task of interest is multimodal in nature. Here, for modality encoding we consider CLIP\footnote{The ViT-L/14 version provided by \url{https://github.com/openai/CLIP}.} \cite{radford2021learning}. After processing the image $g\in\mathbb{R}^{h\times w\times 3}$ and the text $x\in\mathbb{N}^L$, where $h$ and $w$ are the width and height of the image while $L$ is the text’s number of words, with CLIP’s image and text encoders, we get the image embedding $o^g\in\mathbb{R}^d$ and its patch embeddings $\{o_i^g\}_{i=1}^{n_g}\in\mathbb{R}^{n_g\times d}$ as well as the text embedding $o^x\in\mathbb{R}^d$ and its token embeddings $\{o_i^x\}_{i=1}^{n_x}\in\mathbb{R}^{n_x\times d}$, where $n_g$ and $n_x$ are the number of patches and the number of tokens, respectively.

\subsection{External knowledge retrieval and encoding}
Incorporating external knowledge, is inspired by the fact that hate is usually targeted to certain population groups (e.g., muslims, women). Otherwise, the trained models decide purely based on image-text learned correlations. 
Other researchers have realized this issue as well, for instance in \cite{kiela2021hateful} the authors mention:

\say{\textit{[...] Attacking groups perpetrating hate (e.g. terrorist groups) is also not considered hate. This means that hate speech detection also involves possibly subtle world knowledge.}}

We incorporate external knowledge to our model with the following procedure. For each image we get FairFace \cite{karkkainen2021fairface} predictions regarding gender, race and age of all depicted persons (if any) and denote it by $e\in\mathbb{N}^{3\times n_p}$, where $n_p$ is the number of persons and $n_e=3\times n_p$. We then encode this information through a typical embedding layer that is trained along with the rest of the model and produce the corresponding embeddings $\{o_i^e\}_{i=1}^{n_e}\in\mathbb{R}^{n_e\times d}$.

\subsection{Fusion}
We consider a dual-stage modality fusion approach. During stage 1, we produce token-level modality representations that are aware of the level of alignment with the other modality. This kind of fusion has been proposed first in \cite{kumar-nandakumar-2022-hate} but for image- (sentence-) level representations. More precisely, we compute \(f_i^g=o_i^g\otimes o^x\) and \(f_i^x=o_i^x\otimes o^g\), where $\otimes$ denotes element-wise multiplication, to get the fused image and text features, respectively. During stage 2, a Transformer encoder $T(\cdot)$ processes $f_i^g$,$f_i^x$ and $o_i^e$ along with a learnable classification token $CLS$ and produces the corresponding feature representations:
\begin{equation}
    \Big[r_{cls},\{r_i^g\}_{i=1}^{n_g},\{r_i^x\}_{i=1}^{n_x},\{r_i^e\}_{i=1}^{n_e}\Big]=T\Bigg(\Big[CLS,\{f_i^g\}_{i=1}^{n_g},\{f_i^x\}_{i=1}^{n_x},\{o_i^e\}_{i=1}^{n_e}\Big]\Bigg)
\end{equation}

\subsection{Classification}
As classification head for the hatefulness output we consider a typical fully-connected and sigmoid-activated layer of one unit $D(r_{cls})$.

\subsection{Caption supervision}
The vision encoder is likely to learn reduced image features that are advantageous only for the hatefulness detection, ignoring part of the background’s semantics and consequently may overfit. 
To mitigate this potential deviation, we consider an additional supervision signal by reconstructing a description of the background image through a standard image captioning decoder. 
To this end, we first crop the visual part of the memes extracted with Visual Part Utilization (VPU) \cite{koutlis2022memetector} and then consider the Once-For-All (OFA) \cite{cai2020once} generated caption as target. Finally, a Transformer decoder is used to produce the caption based on the fused image features $\{r_i^g\}_{i=1}^{n_g}$.

\section{Experimental setup}\label{sec:experimental}
\subsection{Datasets}\label{subsec:datasets}
\textbf{Facebook Hateful Memes \cite{kiela2021hateful}}: It contains 10K memes labeled as hateful or not and is split in 8.5K training data, 0.5K validation data (dev) and 1K test data.  The test split labels are not released, thus we evaluate all models on the dev set.

\textbf{Memotion7k \cite{sharma2020semeval}}: It contains 9,871 memes, 1K trial data, 6,992 training data and 1,879 test data. It is human annotated for: (a) sentiment prediction, (b) overall emotion prediction and (c) estimation of the corresponding intensities. We randomly sample 10\% from the training set for validation and report results on the test data.

\textbf{MultiOFF \cite{suryawanshi2020multimodal}}: It contains 743 memes manually annotated as offensive or non-offensive. The training set has 445 images and the validation and test sets have 149 each.

\subsection{Baselines}
We consider three baselines, an image-only, a text-only and a multimodal. The image-only is a ResNet18 \cite{He2015DeepRL} pretrained on ImageNet \cite{deng2009imagenet}, the text-only is a trained-from-scratch LSTM \cite{hochreiter1997long}, and the multimodal is a combination through early fusion of the above two models. In addition, we report the top performing competitive methods' scores per benchmark for comparison purposes. Unfortunately, there are no other papers reporting results on all the benchmarks considered by this work, thus the comparison is performed against different methods across the datasets. Brief descriptions about the competitive methods can be found in Section~\ref{sec:related_work}.

\begin{table*}[!ht]
\begin{minipage}[b]{0.32\linewidth}\centering
\begin{tabular}{lll}
\hline
method   & accuracy       & AUC            \\
\hline
image    & 0.530          & 57.3          \\
text     & 0.544          & 62.2          \\
multimodal    & 0.554          & 61.3          \\
\hline
\cite{kiela2020hateful} & 0.659 & 74.14 \\
\cite{muennighoff2020vilio}&-&81.56\\
\cite{lippe2020multimodal} &-&77.39\\
\cite{10.1145/3474085.3475625}&\textbf{0.758}&\textbf{82.8}\\
\cite{Zhou21}&-&73.93\\
\cite{blaier2021caption}&-&78.57\\
\cite{kumar-nandakumar-2022-hate}&-&82.62\\
\hline
MemeFier & 0.736 & 80.1 \\
\hline
\end{tabular}
\caption{Model performance on Facebook Hateful Memes dataset (dev seen) in terms of accuracy and AUC scores.}\label{tab:fbhm}
\end{minipage}
\hspace{0.1cm}
\begin{minipage}[b]{0.32\linewidth}
\centering
\begin{tabular}{llll}
\hline
method     & a              & b              & c 
\\
\hline
image      & 0.333          & 0.502          & 0.315          \\
text       & 0.350          & 0.481          & 0.279          \\
multimodal & 0.346          & 0.493          & 0.310          \\
\hline
\cite{keswani-etal-2020-iitk-semeval} & 0.355 & - & - \\
\cite{vlad-etal-2020-upb}& 0.345&0.518&0.317\\
\cite{guo-etal-2020-guoym}&0.352&0.515&0.323\\
\cite{kumari-etal-2021-co}&0.368&-&-\\
\cite{ouaari2022multimodal}&0.353&-&-\\
\cite{10.1007/978-3-030-99736-6_35}&0.366&0.469&-\\
\cite{10.1007/978-3-030-98358-1_47}&0.370&-&-\\
\hline
MemeFier   & \textbf{0.396} & \textbf{0.519} & \textbf{0.343} \\
\hline
\end{tabular}
\caption{Model performance on Memotion7k dataset in terms of F1 score.}\label{tab:m7k_f1}
\end{minipage}
\hspace{0.1cm}
\begin{minipage}[b]{0.32\linewidth}\centering
\begin{tabular}{lll}
\hline
method     & accuracy       & F1             \\
\hline
image      & 0.638          & 0.619          \\
text       & 0.571          & 0.508          \\
multimodal & 0.671          & 0.626 \\
\hline
\cite{suryawanshi2020multimodal} &-&0.48\\
\cite{10.1145/3474085.3475625}&-&0.646\\
\cite{10.1007/978-3-030-98358-1_47}&-&\textbf{0.671}\\
\hline
MemeFier   & \textbf{0.685} & 0.625 \\
\hline
\end{tabular}
\caption{Model performance on MultiOFF dataset in terms of accuracy and AUC scores.}\label{tab:moff}
\end{minipage}
\end{table*}

\subsection{Hyperparameter tuning}
For the baselines we conduct experiments for hyperparameter tuning accounting for different (1) initial learning rates (1e-2, 1e-3, 1e-4, 1e-5), (2) ResNet18 visual feature extractor being pre-trained on ImageNet or not (True, False), (3) number of hidden dimensions for the fully connected as well as for the LSTM (64, 128, 256), (4) number of LSTM layers (1, 3). We report the maximum performance.

For the proposed method, MemeFier, we consider the following hyperparameter grid. (1) Initial learning rates (1e-4, 1e-5), (2) number of epochs (16, 32), (3) contribution of the caption supervision $\alpha$ (cf. Section~\ref{subsec:implementation}) to the final loss function (0.2, 0.8), (4) model dimension (512, 1024), (5) Transformer encoder settings: (5a) 4 heads, 512 feedforward dimension, 1 layer, (5b) 16 heads, 2048 feedforward dimension, 3 layers, (6) Transformer decoder settings: (6a) 64 input dimension, 4 heads, 64 feedforward dimension, 1 layer, (6b) 256 input dimension, 16 heads, 256 feedforward dimension, 3 layers.

\subsection{Implementation details}\label{subsec:implementation}
For the baselines, training images are resized to 256, randomly cropped at 224, randomly horizontally flipped and standardized using the ImageNet statistics. Validation images are only resized to 224 and standardized. For text, we consider lowercasing and removing punctuation, numbers and double spaces. The vocabulary size is determined by the number of words having at least 5 occurrences in the corpus and the maximum sequence length by the 90\% quantile of the distribution of lengths. Binary cross entropy is employed for the binary classification tasks and the multi-label classification tasks, while categorical cross entropy is employed for the multi-class classification tasks. We use the Adam optimizer, training for 10 epochs, with batch size 128 and the learning rate is reduced by a factor of 10 at epoch 5.

For MemeFier, we consider CLIP's image preprocessing pipeline and the same text preprocessing pipeline used for the baselines. For the text vocabulary we consider the same approach as for the baselines while for the captions vocabulary we consider all words and the actual maximum sequence length. The model comprises almost 29M parameters that we optimize using the Adam optimizer, binary cross entropy loss for the hatefulness output and categorical cross entropy for the caption supervision. Batch size is set to 32 and the learning rate is reduced by a factor of 10 after half epochs. The losses are combined as below:
\[\mathcal{L}=\mathcal{L}_{hate}+\alpha\cdot\mathcal{L}_{caption}\]
$\alpha$ denotes the caption supervision contribution, $\mathcal{L}_{hate}$ denotes binary cross-entropy, and $\mathcal{L}_{caption}$ denotes categorical cross-entropy.

\subsection{Evaluation protocol}
We report F1 score and accuracy for Memotion7k and MultiOFF, and AUC for Facebook Hateful Memes.

\section{Results}\label{sec:results}
Table \ref{tab:fbhm} presents MemeFier performance compared to state-of-the-art on the dev seen split of the Facebook Hateful Memes dataset. In terms of accuracy the multimodal baseline outperforms the two unimodal baselines while in terms of AUC the text based approach performs better than both the image and the multimodal one. State-of-the-art methods exhibit varying performance between 74.1-82.8 in terms of AUC, while MemeFier performs comparably resulting in 80.1 AUC.

Table \ref{tab:m7k_f1} illustrates the performance of MemeFier and competitive methods on the Mememotion7k dataset in terms of macro F1 score. Accuracy is not reported by other papers on this dataset so we do not provide a dedicated table here; however, some marginal gains were observed with MemeFier. Regarding the baselines, we observe that the combination of input modalities does not lead to best results in most of the cases. The latter entails the dominance of one of the two modalities in terms of exploitable information for solving the task of interest and it has already been demonstrated in previous papers as well \cite{das-etal-2020-kafk}. However, MemeFier effectively exploits both modalities and outperforms all baselines and state-of-the-art methods in all tasks (a, b and c) of this dataset.

Table \ref{tab:moff} illustrates the performance of MemeFier and competitive methods on MultiOFF. Image  surpasses text, while the combination of the two modalities leads to the best outcome both in terms of accuracy and F1 score. The proposed method’s performance is better than the baselines in terms of accuracy while almost identical in terms of F1 score. Also, it is comparable but lower than the state of the art.

Finally, Table \ref{tab:ablation} provides an ablation analysis. We ablate the external knowledge input, the caption supervision, the first and the second fusion stage, respectively. We observe that all MemeFier components are necessary to provide the best performance as removing any of them results in reduced accuracy.

\begin{table}[]
    \centering
    \begin{tabular}{lc}
    \hline
        MemeFier & \textbf{80.1} \\
        \hline
        - External knowledge & 78.7 \\
        - Caption supervision & 79.1 \\
        - Fusion stage 1 & 72.8 \\
        - Fusion stage 2 & 67.4 \\
        \hline
    \end{tabular}
    \caption{Ablation analysis of MemeFier on Facebook Hateful Memes dataset. The performance (AUC) of the proposed method after removing (denoted by hyphen) components one by one.}
    \label{tab:ablation}
\end{table}

\section{Conclusions}\label{sec:conclusions}
In this work we propose MemeFier, a deep learning-based architecture to address the task of Internet image meme fine-grained classification. Our method utilizes a dual-stage modality fusion module to first estimate modality alignment at token level via element-wise multiplication and then search for input dependencies at various scales through a typical Transformer encoder. In addition, external knowledge pertinent to protected attributes of the depicted persons is incorporated for joint processing and caption supervision is imposed as a regularization factor. Our pipeline competes and in some cases surpasses state-of-the-art methodologies on three widely-adopted meme classification benchmarks.

\section*{Acknowledgments}
This work is partially funded by the Horizon 2020 European project MediaVerse under grant agreement no. 957252.

\bibliographystyle{unsrt}  
\bibliography{references}

\end{document}